\documentclass[sigconf]{acmart}
\AtBeginDocument{%
  }

\setcopyright{acmlicensed}
\copyrightyear{2026}
\acmYear{2026}
\acmDOI{XXXXXXX.XXXXXXX}
\acmConference[ACM SIGKDD Explorations Newsletter]{Make sure to enter the correct
  conference title from your rights confirmation email}{June 2026}{Woodstock, NY}
\acmISBN{978-1-4503-XXXX-X/2018/06}

\usepackage{multirow}
\usepackage[T1]{fontenc}
\usepackage[utf8]{inputenc}
\usepackage{tabulary}
\usepackage{comment}
\usepackage{enumitem}
\usepackage{microtype}
\usepackage{natbib}
\usepackage{xcolor}

\begin{document}

\title{LLM-Based Scientific Peer Review: Methods, Benchmarks, and Reliability Challenges}
\author{Thi Huyen Nguyen}
\orcid{0000-0001-8195-716X}
\affiliation{%
  \institution{L3S Research Center, Leibniz University Hannover}
  \city{Hannover}
  \country{Germany}
}
\email{nguyen@l3s.de}

\author{Zahra Ahmadi}
\correspondingauthor
\orcid{0000-0003-1110-4756}
\affiliation{%
  \institution{Peter L. Reichertz Institute for Medical Informatics of TU Braunschweig and Hannover Medical School\\Lower Saxony Center for AI and 
Causal Methods in Medicine (CAIMed)}
  \city{Hannover}
  \country{Germany}}
\email{ahmadi.zahra@mh-hannover.de}


\renewcommand{\shortauthors}{Nguyen and Ahmadi}

\begin{abstract}
The rapid growth of scientific submissions has pushed traditional peer review toward its scalability limits, motivating the exploration of large language models (LLMs) as intelligent automated evaluation assistants. Although recent studies show that LLMs can generate fluent critiques and approximate reviewer scores, their reliability, robustness, and security as decision-support systems remain insufficiently understood. This survey offers a systems-level analysis of LLM-based scientific peer review, focusing on two core evaluative functions: critique generation and score prediction. We present a structured taxonomy of modeling approaches (including prompt-based, supervised, retrieval-augmented, and alignment-optimized approaches), and synthesize empirical findings across existing benchmarks. We analyze dataset constraints, evaluation shortcomings,  and domain concentration biases that limit current assessment practices. Beyond performance metrics, we identify emerging robustness risks, including prompt injection, data poisoning, retrieval vulnerabilities, and reward hacking, which expose automated review pipelines to strategic manipulation. From a data mining perspective, we outline key open challenges in modeling subjective disagreement and cross-domain generalization. By reframing automated peer review as a high-stakes, multi-objective decision problem, this survey provides a roadmap for developing robust, transparent, and trustworthy AI-assisted scientific evaluation systems.
\end{abstract}


%\ccsdesc[500]{Do Not Use This Code~Generate the Correct Terms for Your Paper}

\received{23 June 2026}

\maketitle

\section{Introduction}
\label{sec:intro}

\begin{figure}[!t]
    \centering
    \includegraphics[width=\columnwidth]{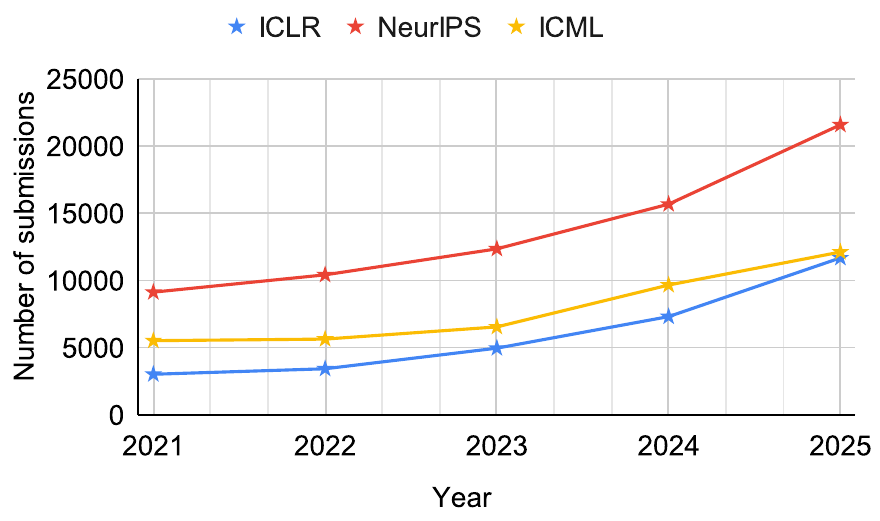}
    \caption{Annual submission counts for three conferences in computer science.}
    \label{fig:subperyear}
\end{figure}

Peer review is the primary quality-control mechanism of scholarly publishing, but its effectiveness increasingly depends on a review workforce that has not scaled with submission volume. Reviewers are expected to provide structured critiques and quantitative recommendations that assess the novelty, soundness, significance, and broader contribution of submitted work. As submission volumes grow, however, this human-centered evaluation process faces mounting pressure from reviewer shortages, subjective judgments, compressed timelines, and limited scalability. 
These pressures are amplified by the rapid growth of scientific output. Scientific publications are estimated to double roughly every decade, whereas the global population of scientists grows by only 21\% over the same period~\cite{kunzli2022not}. Statistics compiled by Paper Copilot~\cite{papercopilot_statistics} show that leading computer science conferences such as ICLR, NeurIPS, and ICML have doubled, or more than doubled, their annual submission counts within five years (2021--2025), as shown in Figure~\ref{fig:subperyear}. As reviewer workload increases, evaluation timelines often shrink, potentially increasing variability in critique depth, score calibration, and review reliability. These challenges have motivated growing interest in AI-assisted peer review systems.

\begin{figure*}
    \centering
    \includegraphics[scale=0.7]{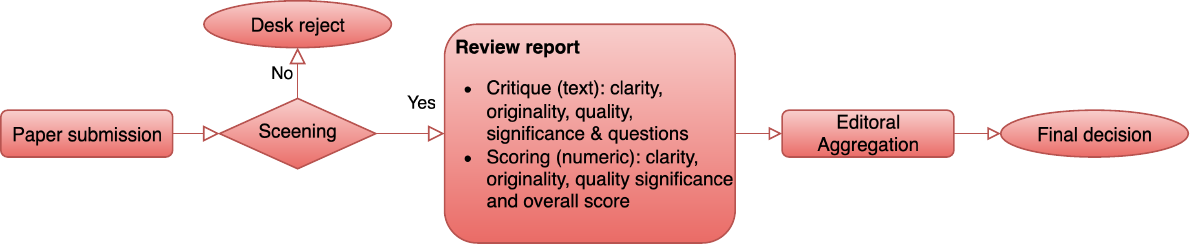}
    \caption{Peer review pipeline.}
    \label{fig:peer_review_pipeline}
\end{figure*}
Early work applied Natural Language Processing (NLP) techniques~\cite{price2017computational,li2019generating,wang2020grammatical,weng2025cycleresearcher} to support or automate specific stages of the review pipeline. With the emergence of LLMs such as GPT-4~\cite{achiam2023gpt}, LLaMA~\cite{touvron2023llama}, and Gemini~\cite{team2023gemini}, automated peer review has shifted from feature-based prediction and limited summarization toward full review generation and score estimation.
Recent studies~\cite{yu2024automated,idahl2025openreviewer} show that LLMs can generate fluent reviewer-like critiques and approximate human scoring patterns. Consequently, LLM-based systems are increasingly being investigated for automating two fundamental components of peer review reports: (1) textual critique generation and (2) quantitative score prediction. These two components are central to editorial and program-committee decision-making. Critiques articulate structured assessments of a manuscript’s strengths and weaknesses, while scores translate these assessments into quantifiable recommendations. Automating these evaluative functions could shift parts of peer review toward a more scalable and data-driven decision pipeline. At the same time, such automation raises fundamental questions about reliability, bias, calibration, and security.

Despite rapidly growing interest in LLM-assisted reviewing, existing surveys~\cite{kuznetsov2024can,zhuang2025large,luo2025llm4sr} do not provide a focused and systematic analysis of automated critique generation and score prediction. Kuznetsov et al.~\cite{kuznetsov2024can} offer a high-level overview of AI assistance across different stages of peer review, spanning activities before, during, and after the evaluation process. Luo et al.~\cite{luo2025llm4sr} survey LLM applications in scientific research more broadly, with peer review discussed briefly as one of many tasks. Zhuang et al.~\cite{zhuang2025large} concentrate more directly on automated scholarly review and relevant datasets, emphasizing the potential of LLMs to alleviate technical bottlenecks. However, these surveys do not provide a structured taxonomy of critique-generation and score-prediction methodologies, nor do they analyze evaluation limitations and robustness concerns in depth.

In contrast, this survey focuses specifically on LLM-based scientific critique generation and score prediction as the core evaluative components of peer review reports. We organize existing work into a structured taxonomy of modeling approaches, synthesize empirical findings across studies, examine dataset and evaluation challenges, and analyze emerging risks in automated scoring pipelines. By treating critique generation and score prediction as decision-critical tasks, we offer a systems-level perspective that complements broader surveys on AI-assisted peer review and LLMs for scientific workflows. Specifically, we aim to answer four key questions: 
\begin{enumerate}[nosep]
    \item How reliable are LLM-based score predictions relative to human reviewers?
    \item To what extent can LLMs generate substantive scientific critiques?
    \item How robust are existing systems under current data and evaluation limitations?
    \item What challenges must be addressed to ensure reliable and secure deployment?
\end{enumerate}

From a modeling perspective, automated peer review can be understood as a high-stakes, multi-objective decision-making problem under noisy supervision. Peer review datasets contain inconsistent decision signals due to inter-reviewer disagreement, acceptance bias, and distribution shift across venues and disciplines. We therefore frame LLM-based peer review not only as a text generation problem, but also as a structured data mining problem involving trade-offs among quality, fairness, calibration, uncertainty, and robustness.

%Peer review datasets exhibit substantial label noise due to inter-reviewer disagreement, class imbalance arising from acceptance bias, and distribution shift across venues and disciplines. Consequently, modeling critique and scoring requires methods for learning under weak or inconsistent labels, estimating uncertainty, calibrating predictions, improving fairness, and generalizing across domains. We therefore frame LLM-based peer review not only as a text generation problem, but also as a structured data mining problem involving noisy decision signals and multi-objective trade-offs among quality, fairness, calibration, and robustness.

% This survey is organized as follows. Section~\ref{sec:peer_review} introduces an overview of the peer review task and its formalizations. Section~\ref{sec:llm_techniques} categorizes current LLM-based review generation approaches. Section~\ref{sec:datasets} presents benchmark datasets and evaluation metrics. Section~\ref{sec:challenges} discusses ethical concerns and challenges. Finally, we outline future directions and opportunities for deployment in Section~\ref{sec:future_direction} and conclude in Section~\ref{sec:conclusion}.

% Section~\ref{sec:applications} summarizes applications of LLMs beyond review generation.

\section{Scientific Peer Review}
\label{sec:peer_review}

The peer review process evaluates scientific manuscripts through assessments by domain experts. %to ensure their quality, originality, and relevance for publication. Widely adopted by journals and conferences, peer review remains a cornerstone of research evaluation. %As illustrated in Figure \ref{fig:peer_review_pipeline}, 
Each submission is typically assigned to one or more reviewers, who evaluate multiple aspects such as clarity, technical correctness, novelty, and potential impact. Reviewers are often guided by venue-specific templates that request a summary, strengths and weaknesses, questions for authors, and a preliminary recommendation, such as acceptance or rejection. High-quality reviews support both editorial decision-making and manuscript improvement; for example, 91\% of researchers report that peer review improved their last publication~\cite{mulligan2013peer}. %re essential not only for selecting impactful research but also for helping authors improve their work:

Scientific peer review can be abstracted as an evaluative decision pipeline, as illustrated in Figure~\ref{fig:peer_review_pipeline}. Although procedures vary across disciplines and publication venues, most peer review systems produce two central outputs: structured textual critique and quantitative score assignment. The textual critique describes the manuscript's strengths and weaknesses across dimensions such as clarity, novelty, technical quality, and significance, while the scoring component converts these qualitative judgments into operational signals, including dimension-specific scores, overall recommendation scores, accept/reject decisions, and reviewer confidence ratings.

From a computational perspective, critique generation and score prediction can be formulated as related modeling tasks. Let $x$ denote a representation of the manuscript. The critique function can be viewed as a conditional generation mapping:
\begin{center}
    $f_c(x) \rightarrow y_{critique}$. 
\end{center}
Meanwhile, the scoring function corresponds to a regression or classification mapping: 
\begin{center}
    $f_s(x) \rightarrow y_{score}$, 
\end{center}
where $y_{critique}$ and $y_{score}$ denote the structured evaluative feedback and numerical evaluation, respectively.

From a machine learning perspective, both functions operate under weak supervision and label uncertainty. Reviewer scores are not deterministic ground-truth labels; rather, they are stochastic realizations influenced by subjective interpretation, reviewer expertise, and venue-specific criteria. Therefore, the target is not a single ``correct" score, but a distribution of plausible judgments shaped by reviewer expertise, venue norms, and uncertainty.
%automated peer review systems must address learning under label noise, uncertainty modeling, and calibration, which are core challenges in data mining.

Peer review, however, differs from standard supervised learning settings in several important respects. First, evaluation is inherently multidimensional. Criteria such as novelty, technical soundness, clarity, and impact may be weighted differently across venues and may evolve over time. Second, inter-reviewer disagreement is common. Multiple reviewers evaluating the same manuscript often produce divergent critiques and assign substantially different scores. This variability raises fundamental questions about the target of prediction: should automated systems predict individual reviewer scores, aggregate scores, or final editorial decisions? Third, the consequences of prediction errors are high. As Figure~\ref{fig:peer_review_pipeline} highlights, critique and scoring play a central role in the decision process. Errors or biases at this stage can propagate directly into editorial aggregation and final publication outcomes. Consequently, analyzing automated systems for critique generation and score prediction requires careful attention not only to generative quality but also to reliability, calibration, robustness, and security. 

These properties make automated peer review substantially more complex than fluent text generation or score approximation alone. Reliable systems must preserve alignment between critique content and numerical recommendations, represent uncertainty arising from reviewer variability, and remain robust under domain shift and potential adversarial manipulation. These structural characteristics motivate the systematic analysis of LLM-based critique generation and score prediction systems developed in the following sections.

%Recently, the rapid growth in submission volume has increased interest in automating peer review report generation. As reviewer workload intensifies, AI-assisted systems capable of generating critiques or calibrating scores are increasingly being explored as decision-support tools. However, automating critique and scoring requires more than generating fluent text or approximating score distributions. Reliable systems must approximate evaluative reasoning, maintain alignment between critique content and numerical recommendations, represent uncertainty arising from reviewer variability, and remain robust under domain shift and potential adversarial manipulation. These structural characteristics motivate a systematic analysis of LLM-based critique generation and score prediction systems, which we examine in the subsequent sections.

\section{LLMs as Reviewers}
\label{sec:llm_techniques}
% \HN{-Focus on different techniques, existing work using LLM for review generation}

\begin{figure}[!t]
    \centering
    \includegraphics[width=\columnwidth, trim={0.86cm 0cm 0cm 1cm},clip]{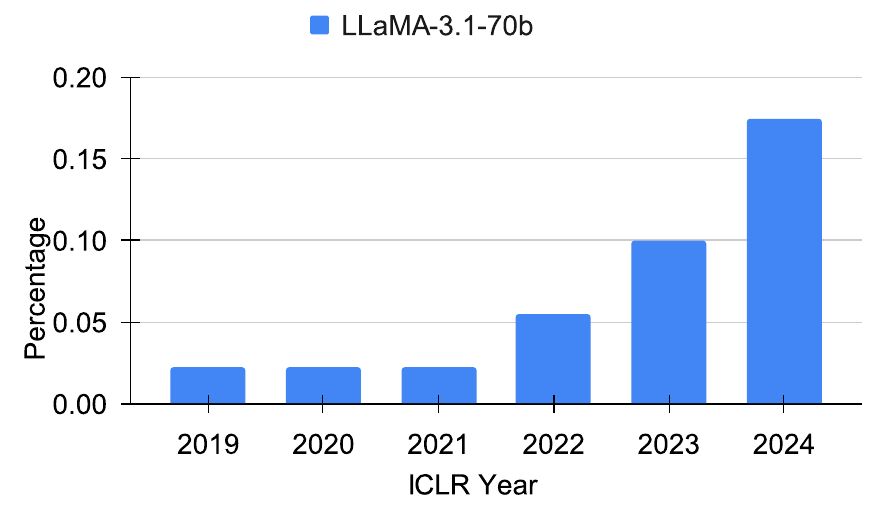}
        \caption{Fraction of reviews detected as LLM-generated by year. }
    % \caption{Percentage of reviews detected as LLM-generated by year\cite{yu2024your}.}
    \label{fig:llm_detected_reviews}
\end{figure}

\begin{figure*}[t]
    \centering
    \includegraphics[width=\textwidth]{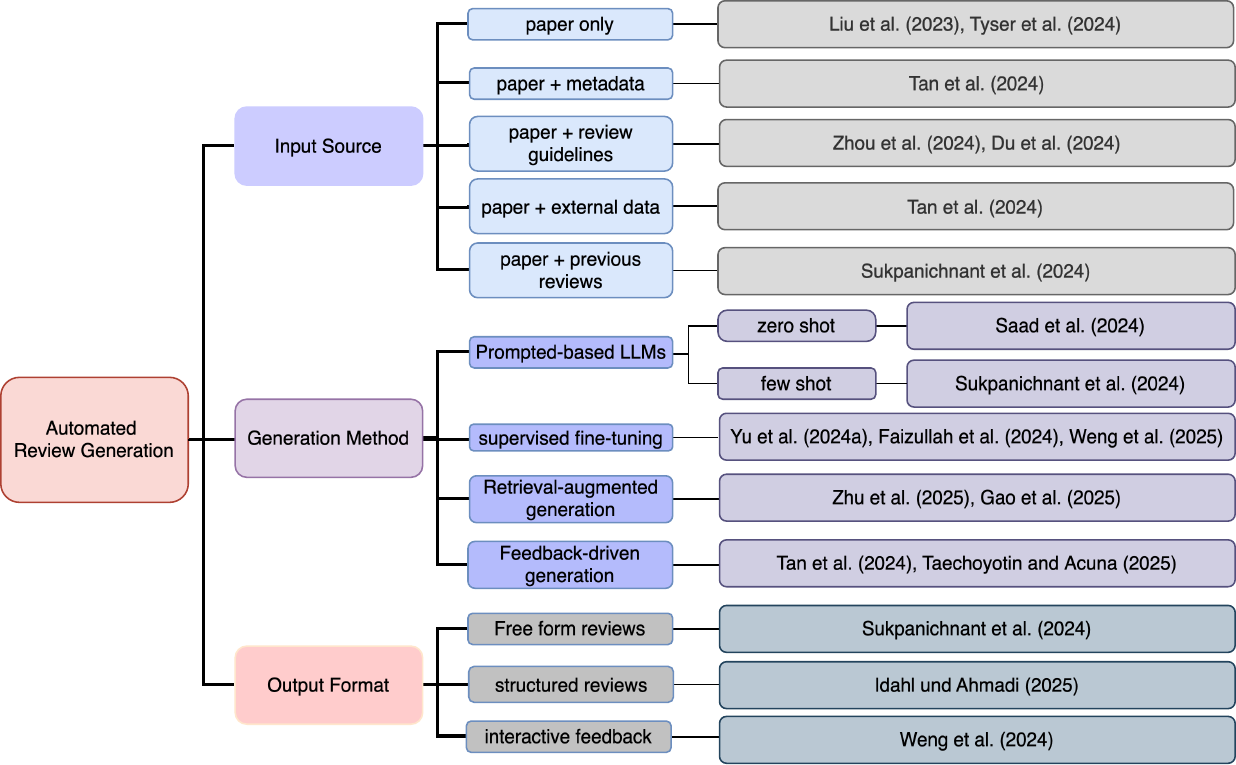}
    \caption{Taxonomy of automated peer review generation, categorized by different aspects.}
    \label{fig:taxonomy_peer_review}
\end{figure*}

Before the emergence of large language models, research on automated peer review primarily focused on score prediction and limited summarization rather than full critique generation~\cite{kang2018dataset,stappen2020uncertainty,dycke2023nlpeer}. Predicting numerical recommendations from paper features was considered more tractable than generating structured evaluative feedback, which requires domain knowledge, contextual understanding, and coherent reasoning. Early supervised approaches~\cite{yuan2022can,lin2023moprd,yuan2022kid} attempted to generate review text directly from manuscript representations, but the resulting reviews were often shallow, fragmented, or template-like. These limitations reflected the difficulty of modeling peer review as judgment-oriented reasoning rather than surface-level text generation.

The rapid advancement of LLMs has substantially shifted this landscape. Recent studies~\cite{latona2024ai,yu2024your} indicate a growing trend in the use of LLMs for review writing. Since the emergence of ChatGPT~\cite{openai2023chatgpt}, the proportion of ICLR reviews flagged as AI-generated has increased sharply~\cite{yu2024your}, as shown in Figure~\ref{fig:llm_detected_reviews}. At least 15.8\% of ICLR 2024 reviews were detected as being written with AI assistance~\cite{latona2024ai}. However, since AI-text detectors are imperfect, these estimates should be interpreted as approximate indicators rather than definitive measurements of AI use.
AI-generated reviews may offer value across multiple dimensions~\cite{tyser2024ai}. For authors, they can provide early, actionable feedback before submission and support manuscript revisions. For reviewers, they may serve as reference material for improving review quality. For journals and conferences, such tools can support quality control and potentially accelerate parts of the peer review workflow. In addition, AI-generated assessments may eventually support reading prioritization, although their reliability for identifying high-quality papers remains uncertain.
%AI-generated review scores may help the academic community identify noteworthy papers, aiding in reading prioritization.

Owing to large-scale pretraining on diverse corpora, LLMs demonstrate strong capabilities in long-form generation, instruction following, and reasoning-style prompting. LLM-based systems often formulate peer review as a general text-generation task: Given a manuscript, generate a complete review report. In many such systems, critique generation and score prediction are not explicitly modeled as separate functions. Instead, models directly produce free-form review text that may implicitly include strengths, weaknesses, and recommendations. 

A growing body of work has examined the capabilities and limitations of LLM-based review report generation. Figure~\ref{fig:taxonomy_peer_review} provides a taxonomy of existing work, categorized along several dimensions. Existing approaches can be broadly grouped into four paradigms: prompt-based systems, fine-tuned systems, retrieval-augmented systems, and alignment-optimized systems.

\subsection{Prompt-based LLMs}
%Prompt-based learning has emerged as a key strategy for adapting LLMs to downstream tasks without fine-tuning. 
Prompt-based systems are the most accessible form of LLM-assisted reviewing, but their flexibility comes at the cost of sensitivity to instructions and weak score calibration.
In automated peer review generation, prompt-based approaches allow general-purpose LLMs, such as GPT-4~\cite{achiam2023gpt}, LLaMA~\cite{touvron2023llama}, and Gemini~\cite{team2023gemini} to be guided by carefully constructed instructions that simulate human reviewer behavior. Rather than explicitly training models on peer review datasets, these systems rely on instruction-following capabilities acquired during large-scale pretraining. Prompts typically ask the model to assess novelty, technical soundness, clarity, strengths and weaknesses, and to provide an overall recommendation score.

The appeal of prompt-based approaches lies in their flexibility and scalability~\cite{liu2023pre}. Generating high-quality reviews goes far beyond simple summarization; it requires critiques and evaluations across multiple dimensions, including novelty, significance, methodology, and clarity. Prompting enables LLMs to generate outputs that follow predefined templates, criteria, or word limits, making such approaches particularly attractive when large annotated review datasets are unavailable. This low-cost adaptability has driven widespread experimentation with prompt-based peer review systems.

\textbf{Prompting strategies}. Two main prompting strategies are commonly used: zero-shot prompting ~\cite{kojima2022large}, where the model receives a single instruction and relies on pretraining to generate an output, and few-shot prompting~\cite{min2022rethinking}, where a small set of input-output examples is provided to guide the model's response. Depending on their design, prompts for LLM-based peer review can take several forms, such as criterion-based, section-based, style-guided, review-score paired, or chain-of-thought formats. 

Beyond these basic paradigms, prompt designs for peer review tasks vary considerably:
\begin{itemize}[nosep]
    \item Criterion-based: Explicitly request evaluation across multiple dimensions.
    \item Style-guided: Instruct the model to follow venue-specific tone or templates.
    \item Review–score paired: Require both textual feedback and numerical ratings.
    \item Section-based: Focus on specific manuscript components.
    \item Chain-of-thought: Encourage step-by-step reasoning before final judgment.
\end{itemize}
Table~\ref{tab:prompt_based_LLMs} illustrates different prompt designs, examples, strengths, and limitations. These designs implicitly shape how the model approximates the critique function $f_c(x)$ and, when applicable, the scoring function $f_s(x)$. For example, review–score paired prompts attempt to jointly elicit textual reasoning and quantitative recommendations, whereas section-based prompts decouple local evaluation from global assessment.

\begin{table*}[!t]
    \centering
    \scriptsize
     \renewcommand{\arraystretch}{1.4}
     \setlength{\tabcolsep}{0.1em}
    \begin{tabulary}{\textwidth}{p{1.2cm}|>{\centering\arraybackslash}p{4cm}|>{\centering\arraybackslash}p{6cm}|>{\centering\arraybackslash}p{2.5cm}|>{\centering\arraybackslash}p{3.7cm}}
    \hline 
    Prompt type & Description & Prompt Example & Strength & Limitation\\
    \hline 
     criterion-based \cite{markhasin2025ai}    & Prompts LLMs to write a review across multiple dimensions such as novelty, clarity, etc.  & Evaluate the given paper on aspects: novelty, significance, clarity, and quality& Generate structured reviews & May lead to shallow, checklist-style reviews\\

     \hline
      style-guided \cite{liang2024can}   & Instructs LLMs to follow specific tone, review template, or style& Write a constructive, formal review following ICLR guidelines& Generate context-appropriate reviews& May constrain creativity or critical depth, reduce content diversity or critical honesty.\\
      \hline 
      review-score pair \cite{saad2024exploring}&Requires both numeric rating and review comments & Give a review and score evaluation on the following aspects: novelty, clarity, significance, quality&Simulate human-like reviews &  Inconsistent score-text alignment.\\
      \hline 
      section-based \cite{d2024marg}&Focuses prompts on paper sections (e.g., abstracts, methodology)&Review methodology section for strengths and limitations& Enable fine-grained evaluation & Misses global context of the paper\\ 
      \hline 
      chain-of-thought \cite{stahl2024exploring}& Instructs the model to reason step-by-step & Step 1: what is the paper's main claim? Step 2: is the claim supported? Step 3: evaluate the strengths and weaknesses of the methodology& Improve factuality, coherence, and reasoning depth & Increase output length, slower generation.\\
      \hline
    \end{tabulary}
    \caption{Prompt-based approaches categorized by prompt design.}
    \label{tab:prompt_based_LLMs}
\end{table*}

\textbf{Findings}. Early studies~\cite{robertson2023gpt4,liang2024can} provided empirical evidence that GPT-4, even with zero-shot prompting, can meaningfully contribute to the peer review process. Similarly, Biswas \textit{et al.} \cite{biswas2023chatgpt} found that ChatGPT can generate consistent evaluations and helpful feedback, but it struggles with contextual understanding and subjective interpretation. Other work has assessed ChatGPT-3.5 and ChatGPT-4 using zero-shot prompts~\cite{saad2024exploring}, finding that their outputs were only weakly correlated with final acceptance decisions and tended to be overly positive. 

Few-shot approaches have shown improvements in structural alignment and acceptance prediction~\cite{sukpanichnant2024peerarg}. However, fine-grained analyses~\cite{du2024llms} indicate that while LLMs produce coherent summaries and high-level evaluations, they frequently miss subtle methodological weaknesses or nuanced experimental limitations. Similarly, evaluations~\cite{zhou2024llm} across GPT-3.5 and GPT-4 reveal persistent challenges in long-document processing, zero-shot score calibration, and producing critiques that match human reviewers in critical depth.

Multi-agent prompting strategies~\cite{d2024marg}, in which different agents focus on clarity, methodology, experiments, or impact, attempt to enhance specificity and reduce generic feedback. Although such approaches improve aspect coverage, they also increase system complexity and do not fully resolve the limitations of prompt-based review generation.

The performance characteristics of prompt-based systems can be understood through their reliance on pretrained patterns. LLMs approximate critique generation by mapping manuscript content into familiar evaluative language structures encountered during training. This explains their fluency and structural coherence. However, because they are not explicitly trained to assess experimental validity or methodological correctness, models may reward clear structure and polished prose while overemphasizing stylistic signals relative to substantive contribution. This behavior reflects both pretraining distribution biases and alignment objectives that favor constructive tone. Effective prompt engineering is therefore critical for improving review structure and critical depth~\cite{markhasin2025ai}. Techniques such as multi-step prompting and chain-of-thought reasoning~\cite{wei2022chain} have been shown to better simulate human reviewer reasoning and improve the factual accuracy.
 
\textbf{Structural limitations}. Beyond empirical performance metrics, prompt-based systems exhibit structural vulnerabilities that affect reliability:
\begin{itemize}[nosep]
   \item Prompt sensitivity: Small changes in instruction wording can significantly alter critique tone and numerical outputs, undermining reproducibility~\cite{zhao2021calibrate}.
   \item Hallucination: Models may misinterpret manuscript content and generate factually inaccurate outputs~\cite{huang2025survey}.
   \item Score–-text inconsistency: Jointly generated critiques and scores may lack logical alignment when numerical outputs are not explicitly constrained~\cite{bharti2026co}.
\end{itemize}

Overall, prompt-based systems are best understood as strong review-format generators rather than reliable evaluators: they can reproduce the style and structure of reviews, but their judgments remain sensitive to prompting, calibration errors, and unsupported claims.
%Overall, prompt-based systems achieve impressive linguistic fluency, but their evaluative reliability remains constrained by pretraining-induced biases.  %while retrieval augmented generation~\cite{lewis2020retrieval} could be explored to reduce hallucination.

\subsection{Supervised fine-tuning}
Supervised fine-tuning~\cite{gao2024reviewer2,idahl2025openreviewer} improves domain alignment by learning from historical reviews, but it also inherits the noise, bias, and subjectivity of peer-review data.
In this paradigm, LLMs are trained on datasets that pair input manuscripts with expert-written peer reviews, enabling models to learn mappings from paper content to review text through instruction-tuned language modeling. 
The objective is to capture a reviewer evaluation style, adherence to review criteria, and structured reasoning patterns. 

\textbf{Fine-tuning strategies}. Supervised fine-tuning can be implemented through two main adaptation strategies: full fine-tuning and parameter-efficient fine-tuning (PEFT). In full fine-tuning~\cite{idahl2025openreviewer}, all model parameters are updated using peer review supervision, enabling maximal adaptation to domain-specific evaluation patterns. However, full fine-tuning requires substantial computational resources and increases the risk of overfitting, particularly when review datasets are limited in size.

Alternatively, PEFT methods such as LoRA~\cite{hu2022lora} introduce a small number of trainable low-rank adaptation parameters while keeping the base model largely frozen. PEFT approaches substantially reduce memory and compute requirements, making them practical for training on modest-sized peer review datasets. Several recent peer review generation systems~\cite{gao2024reviewer2,faizullah2024limgen} adopt LoRA-style adaptation to balance domain alignment with computational feasibility. Although these strategies differ in efficiency and adaptation capacity, both remain fundamentally constrained by the quality and representativeness of available review data.

Depending on the training objective, models can be optimized to generate criterion-specific evaluations, predict scores, or produce complete review reports. Compared to prompt-based systems, supervised fine-tuning explicitly approximates the critique function $f_c(x)$ and, when applicable, the scoring function $f_s(x)$ using labeled review data.

\textbf{Findings}. Recent work demonstrates that supervised fine-tuning improves structural fidelity and domain alignment. For example, Yu \textit{et al.}~\cite{yu-etal-2024-automated} introduced SEA, a fine-tuning framework with three modules: standardization, evaluation, and analysis. Their results show that SEA can produce feedback closely aligned with human reviews. 

Several studies have also focused on specialized subtasks. Faizullah \textit{et al.}~\cite{faizullah2024limgen} 
evaluated PEFT-based models for generating suggestive limitations for a given research paper and demonstrated improvements in specificity over prompt-only approaches.   
Weng \textit{et al.}~\cite{weng2025cycleresearcher} curated 5K reviews and proposed a closed-loop research–review–revision cycle, powered by iterative preference training. Zhu \textit{et al.}~\cite{zhu2025deepreview} constructed a 13K annotated dataset capturing intermediate reasoning steps, enabling multi-stage training for structured review generation. Idal and Ahmadi~\cite{idahl2025openreviewer} constructed a large-scale dataset of 79K reviews from OpenReview and developed a fully fine-tuned LLaMA-based model that generated more critical and realistic reviews than zero-shot baselines. Moreover, Mostafa \textit{et al.}~\cite{mostafa2026novel} fine-tuned a LLaMA-based model to generate human-like novelty assessments and calibrated novelty scores.

Across these studies, fine-tuned models typically outperform prompt-based systems in several aspects:
\begin{itemize}[nosep]
\item Review structure fidelity
\item Criterion coverage
\item Domain-specific terminology usage
\item Correlation with reviewer scores
\end{itemize}

\textbf{Structural limitations}. Despite their empirical advantages, supervised fine-tuned systems face structural constraints:
\begin{itemize}[nosep]
    \item Label noise and reviewer disagreement: Human review data contain substantial inter-reviewer variability. Treating individual scores as ground truth risks amplifying subjective bias.
    \item Dataset concentration: Because most training data come from open-review computer science venues, fine-tuned models may learn venue-specific reviewing norms rather than general principles of scientific evaluation.
    %Most publicly available review datasets originate from computer science venues with open review policies. Limited disciplinary diversity restricts generalization.
    \item Data imbalance: Accepted papers are often overrepresented relative to rejected papers, leading to skewed score distributions and reduced discrimination power.
    \item Bias amplification: Fine-tuned models may encode historical inequities or stylistic preferences embedded in training data.
    \item Ethical and legal constraints: Reviewer anonymity, consent, and data ownership issues complicate large-scale dataset construction and redistribution.
\end{itemize}

Overall, supervised fine-tuning is a promising direction for enhancing the realism and domain relevance of AI-assisted peer reviews. Yet its reliability remains bounded by the quality and representativeness of available datasets. 

%\cite{tan2024peer} constructed a dataset containing over 26,841 papers with 92,017 reviews. Their experiments showed superior performance of fine-tuned LLMs compare to zero-shot settings.

\subsection{Retrieval-augmented generation}
Retrieval-augmented systems address grounding and novelty assessment, but shift the reliability bottleneck from generation to evidence selection. LLMs, even when fine-tuned, may hallucinate unsupported claims, misinterpret experimental details, or generate generic critiques detached from manuscript-specific evidence. To mitigate this issue,  retrieval-augmented generation (RAG) frameworks~\cite{gao2023retrieval} have been explored, which couple LLMs with external retrieval systems to enhance factual consistency, citation awareness, and review depth.

In RAG-based approaches, models are provided with additional evidence beyond the raw manuscript text. Retrieved context may include specific sections of the paper, cited references, related work from external corpora, or structured representations such as knowledge graphs. LLMs then condition their critique and scoring generation on this retrieved information. Conceptually, RAG modifies the critique function $f_c(x)$ by enriching the input representation with auxiliary evidence 
$r(x)$, yielding $f_c(x, r(x))$, with the goal of improving grounding and reducing hallucination.
In peer review, retrieval can serve several functions:
\begin{itemize}[nosep]
\item Intra-document grounding: Selecting relevant manuscript sections for focused critique, such as experimental details.
\item Inter-document comparison: Retrieving related work to assess novelty and contribution.
\item Citation verification: Validating references and claims against external literature.
\item Context enrichment: Providing background information for domain-specific terminology.
\end{itemize}

\textbf{Findings}. Early work, such as ReviewRobot~\cite{wang-etal-2020-reviewrobot}, built knowledge graphs from the target manuscript, its cited works, and background literature to generate structured, evidence-backed critiques and predict review scores. By explicitly modeling relationships among contributions and prior work, ReviewRobot aimed to support more informed novelty and relevance assessment. Similarly, Mostafa \textit{et al.}~\cite{mostafa2026novel} demonstrated that integrating a literature-aware retrieval component improves human-like critiques and calibrated novelty scores with
interpretable, grounded justifications. 

Several recent frameworks incorporate retrieval into multi-stage reasoning pipelines. Zhu \textit{et al.}~\cite{zhu2025deepreview} integrated structured literature search into its generation process, while Gao \textit{et al.}~\cite{gao2025reviewagents} retrieved relevant publications to validate novelty claims and strengthen evaluative grounding. Empirical evaluations in these works suggest that retrieval improves critique specificity and reduces overly generic feedback compared to prompt-only baselines.

Retrieval augmentation addresses a core weakness of LLM-based reviewing: reliance on parametric knowledge encoded during training. Without retrieval, models must infer novelty and methodological validity from internalized statistical patterns, which may be outdated or incomplete. By incorporating external evidence, RAG reduces dependence on memorized knowledge and encourages context-aware reasoning. RAG systems can therefore improve factual alignment and reduce unsupported claims. In scoring tasks, retrieval may also enhance novelty estimation and contribution assessment.

\textbf{Structural limitations}. Despite its promise, retrieval-augmented review generation introduces new challenges:
\begin{itemize}[nosep]
    \item Retrieval bottlenecks: Identifying the most relevant segments from long manuscripts remains difficult due to token-length constraints and segmentation choices.
    \item Corpus dependency: Performance depends heavily on the completeness and quality of the retrieval corpus, such as arXiv or PubMed.
    \item Error propagation: Retrieval errors directly influence generation and may amplify inaccuracies.
    \item Computational overhead: Indexing and querying large scientific corpora increase system complexity and latency.
    \item Confidentiality concerns: Using external APIs or third-party corpora may conflict with blind review protocols.
\end{itemize}

In summary, retrieval-augmented approaches can improve evidence-grounded critique generation and novelty assessment in automated peer review. By coupling LLMs with structured retrieval mechanisms, these systems aim to reduce hallucination and enhance contextual reasoning. However, retrieval-augmented systems shift the bottleneck of prompt-based or fine-tuned methods from language generation to information selection. Their effectiveness depends not only on LLM capacity, but also on retrieval design, corpus coverage, and evidence integration strategies. Practical deployment therefore requires careful attention to retrieval quality, computational efficiency, and confidentiality constraints. 

\subsection{Feedback-driven approaches}

While prompt-based, fine-tuned, and retrieval-augmented systems primarily operate in a static generation paradigm, feedback-driven approaches introduce iterative refinement mechanisms into automated peer review generation. These systems incorporate human or programmatic feedback signals to improve critique quality, structure, and alignment over time. Rather than relying solely on pretrained knowledge or supervised paper-–review pairs, feedback-driven models attempt to approximate reviewer behavior through interaction histories, preference judgments, or reinforcement objectives. Feedback may originate from reviewers, editors, authors, or automated quality checks.

Conceptually, these approaches extend the critique and scoring functions $f_c(x)$ and $f_s(x)$ by incorporating feedback signals $h$, yielding adaptive mappings 
$f_c(x, h)$ and $f_s(x, h)$. The goal is to improve alignment with human evaluative standards beyond what static training can achieve.

\begin{table*}[t]
\resizebox{\textwidth}{!}{
\setlength{\tabcolsep}{0.0em}
    \begin{tabular}{c|c|c|c}
    \hline 
        Paradigm & Main strength & Main limitation & KDD challenge  \\
    \hline
    \centering
Prompt-based &Flexible, low-cost, structured outputs & Prompt sensitivity, hallucination, weak calibration & Robust prompting and uncertainty estimation \\
Fine-tuned & Better domain alignment and review style & Inherits noisy labels and venue bias   & Learning from subjective and biased supervision \\
RAG-based & Better grounding and novelty assessment & Retrieval errors and corpus bias & Reliable evidence retrieval and corpus coverage \\
Feedback-driven & More helpful and aligned critiques & Reward hacking and preference bias & Multi-objective optimization and evaluation\\
\hline 
    \end{tabular}
    }
    \caption{Comparative synthesis of LLM-based peer review paradigms, highlighting their main strengths, structural limitations, and associated data mining challenges.}
    \label{tab:paradims}
\end{table*}

\textbf{Findings}. Recent work has begun to model multi-round review dynamics explicitly. For instance, Tan \textit{et al.}~\cite{tan2024peer} constructed a multi-turn dialogue dataset including reviewer comments, author rebuttals, reviewer responses to rebuttals, and final decisions. These reviewer-author interaction signals, such as rebuttals or revision histories, are used to refine review generation and enable more adaptive and context-aware critiques. In parallel, reinforcement learning from human feedback (RLHF) has been explored to align review generation with expert preferences. Taechoyotin \textit{et al.}~\cite{taechoyotin2025remor} integrated reasoning-enhanced fine-tuning with multi-objective reinforcement learning guided by human-aligned reward functions. Their empirical results suggest improvements in perceived helpfulness, reasoning depth, and structural coherence relative to supervised baselines.

Across these studies, feedback-driven approaches generally improve constructiveness and reduce overly generic outputs. Compared to purely supervised models, alignment-optimized systems tend to produce critiques judged by human evaluators as more actionable and balanced. By incorporating feedback signals, such as preference comparisons, helpfulness ratings, or rebuttal responses, models can be optimized to better match desired evaluation properties. 

\textbf{Structural limitations}. Despite their promise, feedback-driven systems face significant challenges:
\begin{itemize}[nosep]
    \item Reward specification: Designing reward functions that capture true evaluative rigor is difficult. Over-optimization for helpfulness or politeness may reinforce optimism bias.
    \item Preference bias: Human feedback reflects subjective preferences that may vary across disciplines or reviewer communities.
    \item Data scarcity: Collecting high-quality preference annotations or multi-round review interactions is resource-intensive.
    \item Over-optimization risk: Reinforcement learning may produce outputs that maximize reward signals without improving factual correctness.
    \item Calibration neglect: Alignment objectives often prioritize tone and coherence rather than uncertainty modeling or score calibration.
\end{itemize}

Overall, feedback-driven and reinforcement-based approaches represent an important evolution in automated peer review generation. Compared with prompt-based systems, feedback-driven models introduce adaptive refinement mechanisms that enhance constructiveness and structural coherence. Compared to supervised fine-tuning, RLHF-style alignment shifts the objective from historical replication to preference optimization. By incorporating human preferences, multi-turn interactions, or reward-guided optimization, feedback-driven systems can produce more constructive, context-aware, and expert-aligned critiques. However, their effectiveness depends critically on the quality of feedback signals and reward design.

\subsection{Synthesis of empirical patterns}
Beyond surface performance comparisons, automated peer review systems should be evaluated through a multi-objective lens. Effective systems must jointly optimize critique quality, score calibration, fairness across domains and institutions, robustness to adversarial manipulation, and uncertainty awareness. These objectives are often in tension. For example, optimizing for helpfulness may reduce critical sharpness, whereas maximizing score correlation may amplify historical biases embedded in training data. 
In this section, we synthesize empirical evidence across studies to characterize the current performance landscape of automated peer review systems.
Table~\ref{tab:paradims} summarizes the four major LLM-based peer review paradigms discussed above, highlighting their primary strengths, structural limitations, and corresponding data mining challenges. Building on this comparison, the following synthesis examines two recurring empirical patterns across these paradigms: critique quality and score prediction reliability.

\textbf{Critique quality}. Across studies, the main distinction is between surface-level review quality and evaluative rigor: LLM-generated reviews are often readable and well-structured, but their ability to detect subtle methodological flaws remains limited. 
%Across studies, a consistent finding is that LLM-generated critiques achieve high linguistic fluency and structural coherence. 
Human evaluators frequently judge such outputs as readable and well organized~\cite{biswas2023chatgpt,robertson2023gpt4,liang2024can}. However, deeper analyses reveal a gap between surface coherence and evaluative rigor. Fine-grained comparisons with human-written reviews indicate that LLM critiques often emphasize high-level summaries and general improvement suggestions while underrepresenting subtle methodological weaknesses~\cite{du2024llms,yu2024your}. Existing studies show that LLM-generated comments tend to be more positive and contain fewer explicit weakness-oriented statements than human reviews, particularly when identifying limitations or experimental flaws. Moreover, LLMs often
produce paper-unspecific reviews without detailed justification. They remain
prone to hallucination, generating outputs that are plausible-sounding
but factually inaccurate or unverified~\cite{achiam2023gpt,ji2023survey}. Similarly, some studies~\cite{zhou2024llm,chen2025envisioning} report that while GPT-4 produces coherent commentaries, its ability to generate critiques matching expert depth remains limited, especially for long and technically dense manuscripts.

\textbf{Score prediction}. 
Score prediction performance is commonly measured through correlation with reviewer-evaluated scores or final acceptance outcomes. Studies evaluating GPT-3.5 and GPT-4 report weak to moderate correlations with final decisions~\cite{saad2024exploring,zhou2024llm}. Several studies also observe that predicted scores cluster around mid-range categories and show weaker discrimination between accepted and rejected papers than human reviewers~\cite{saad2024exploring,zhou2024llm}. Fine-tuned systems trained on OpenReview datasets often improve alignment with human-evaluated scores~\cite{idahl2025openreviewer}, although the magnitude of improvement varies across datasets. 
Across reported studies, correlations between LLM-generated scores and human reviewer scores typically fall within weak-to-moderate ranges, with fine-tuned models consistently outperforming zero-shot prompting baselines. Retrieval augmentation tends to improve critique specificity and novelty assessment, although gains in score calibration remain limited. Importantly, the achievable upper bound is constrained by intrinsic inter-reviewer disagreement documented in peer review literature~\cite{bornmann2011scientific,pier2018low}, suggesting that perfect alignment with any single reviewer is neither realistic nor desirable. These observations suggest that score prediction should be evaluated distributionally: models should be tested on calibration, uncertainty intervals, rank stability, and agreement with reviewer-score distributions, not only on correlation with average scores.
%These observations indicate that future evaluation should move beyond raw correlation and incorporate calibration metrics, uncertainty estimation, and robustness under domain shift.

% \textbf{Bias and Feature Sentivity}. Empirical analyses also reveal sensitivity to stylistic and surface-level features. Models tend to assign more favorable evaluations to well-written and clearly structured manuscripts, potentially overemphasizing clarity relative to methodological strength~\cite{biswas2023chatgpt}. This pattern mirrors observations in human peer review studies but may be amplified by pretrained language priors.

%However, score correlation must be interpreted cautiously. Reviewer disagreement is well-documented in peer review literature~\cite{bornmann2011scientific,pier2018low}, and variance among human evaluators constrains the upper bound of achievable agreement. A model that approximates average reviewer behavior may achieve moderate correlation without demonstrating strong evaluative reliability. Hence, the reliability and interpretability of automated scoring remain open questions.

% \textbf{Cross-Domain Evidence Gaps}.
% The majority of studies focus on computer science venues with open review policies. Few studies systematically evaluate models on biomedical, social science, or humanities peer review data. Given differences in evaluation criteria across fields, it remains unclear whether current LLM-based systems generalize beyond CS/ML contexts. The lack of cross-domain benchmarks restricts conclusions regarding universal applicability.
\section{Benchmark Gaps and Evaluation Challenges} 
\label{sec:datasets}
While emerging datasets and evaluation protocols have supported recent progress in LLM-based peer review generation, existing benchmarks remain fragmented and limited. Current resources provide valuable starting points, but they fall short of enabling rigorous, cross-domain, and reliability-focused evaluation. In this section, we examine structural gaps in available datasets and highlight methodological challenges in evaluating automated critique and scoring systems.
\subsection{Datasets}

\begin{table*}
    \centering
    \footnotesize
     \resizebox{0.95\textwidth}{!}{
    \renewcommand{\arraystretch}{1.2}
     \setlength{\tabcolsep}{0.0em}
    \begin{tabular}{c|c|c|c}
    \hline 
        Name & Data Source & Data Size & Application \\
    \hline
      PeerRead  &ICLR 2017& 14K papers + decision& score prediction\\ 
      ~\cite{kang2018dataset} &ACL 2017&(3K papers + 10.7K reviews; &acceptance prediction\\
       &NeurIPS 2013-2017&1.3K papers + aspect scores)&\\
    \hline 
    Interspeech&Interspeech 2019&2.1K papers&acceptance prediction, score prediction\\
    ~\cite{stappen2020uncertainty}&&5.8K reviews + decision&\\
    \hline 
    PeerAssist&ICLR 2017-2020&4.4K papers+ 13.4K reviews&acceptance prediction\\
    \cite{bharti2021peerassist}&&&\\
    \hline 
    ASAP-Review&ICLR 2017-2020,
&8.8K papers + 28.1K reviews&review report generation\\
~\cite{yuan2022can}&NeurIPS 2016-2019&&\\
    \hline 
    NLPEER&ACL 2017, ARR 2022,&5.6K papers + 11.5K reviews&score prediction, pragmatic labelling\\
    ~\cite{dycke2023nlpeer}&COLING 2022,
CONLL 2016&&guided skimming for peer review\\
    \hline 
    MOPRD&PeerJ journals&6.5K papers +  22.4K reviews&review report generation, meta-review generation,\\
    ~\cite{lin2023moprd} &&11.2K rebuttal letters&acceptance prediction, rebuttal generation,\\
    &&&scientometric analysis\\
    \hline 
    AgentReview&ICLR 2020-2023&500 papers &review report generation\\
   \cite{jin2024agentreview} &&10.4K reviews + rebuttals&meta-review generation\\
    % \hline 
    % Reviewer2&ICLR 2017-2023&&\\
    % \cite{gao2024reviewer2}&NeurIPS 2016-2022&&\\
    % &PeerRead, NLPEER&&\\
    \hline 
    ReviewMT&ICLR 2017-2024& 26.8K papers + 92.0K reviews&review report generation\\
    \cite{tan2024peer}&&&\\
    \hline
    % ReviewCritique~\cite{du2024llms}&NLP&100 papers + 380 reviews& evaluation\\

    % \hline 
    % AgentReview~\cite{jin2024agentreview}&NLP, ML&500 papers&score prediction, acceptance prediction&\\
    
    % &&10,4K reviews+rebuttals&review report generation&\\
    % &&9.1K meta-review + decision&meta-review geneation&\\
    % \hline
     Review-5K~\cite{weng2025cycleresearcher}&ICLR 2024&5K papers + 16K reviews&review report generation, score prediction\\
 
    \hline 
    DeepReview-13K&ICLR 2024-2025&13.3K papers + reviews&score prediction, acceptance prediction\\
   ~\cite{zhu2025deepreview} &&&review report generation\\
   
    \hline
    PeerRT&ICLR 2017-2020 &5.5K papers + 16.8K reviews&score prediction, review report generation\\
    ~\cite{taechoyotin2025remor}&&&\\
\hline 
OpenReviewer&ICLR 2022-2024&36K papers + 79K reviews&review report generation\\
\cite{idahl2025openreviewer}&NeurIPS 2022-2024&&score prediction\\

\hline 
$Re^2$~\cite{zhang2025re}&45 venues
&19.9K papers + 70.6K reviews&score prediction, review report generation\\
&2017-2025&&acceptance prediction\\
\hline 
    \end{tabular}
    }
    \caption{Datasets, with code for downloading or crawling when available, for automated peer review generation and related evaluation tasks.}
    \label{tab:dataset}
\end{table*}

% Apart from datasets listed in Table~\ref{tab:dataset}, recent works~\cite{gao2024reviewer2,tan2024peer,idahl2025openreviewer} have drawn their own data from the well-known OpenReview~\footnote{openreview.net} platform. 

Many studies have developed datasets to support peer review modeling, and these resources now serve as key benchmarks for measuring and comparing model performance. Table~\ref{tab:dataset} summarizes representative datasets for automated review report generation, score prediction, and acceptance prediction tasks. Most datasets are derived from open review platforms such as OpenReview~\footnote{\url{https://openreview.net}}, PeerJ\footnote{\url{https://peerj.com}}, and F1000Research\footnote{\url{https://f1000research.com}}. They include early benchmarks such as PeerRead~\cite{kang2018dataset}, NLPEER~\cite{dycke2023nlpeer}, and MOPRD~\cite{lin2023moprd}, as well as more recent large-scale collections such as ReviewMT~\cite{tan2024peer}, DeepReview-13K~\cite{zhu2025deepreview}, OpenReviewer~\cite{idahl2025openreviewer}, and $Re^2$~\cite{zhang2025re}.
Despite growing dataset sizes, several structural limitations persist.

\textbf{Data scarcity and confidentiality}.
Peer review is inherently confidential. Most journals and conferences operate under closed-review policies, limiting public access to data. Existing datasets therefore rely primarily on opt-in mechanisms or open-review venues. Consequently, even the largest datasets remain small relative to typical LLM pretraining corpora. Early attempts to expand dataset coverage, such as matching arXiv submissions to later conference proceedings~\cite{kang2018dataset}, provide acceptance labels but not full review texts, thereby limiting their utility for training critique-generation models.
Thus, current benchmarks remain constrained in both scale and representativeness.

\textbf{Domain concentration and distribution bias}. A prominent pattern across Table~\ref{tab:dataset} is the dominance of computer science (CS) venues, particularly machine learning and natural language processing conferences. Although datasets such as MOPRD~\cite{lin2023moprd} and NLPEER~\cite{dycke2023nlpeer} include some multidisciplinary content, their scale remains insufficient for robust cross-domain modeling. This concentration creates two risks. First, apparent performance improvements may reflect alignment between model pretraining data and CS-specific evaluation norms. Second, generalization to disciplines such as biomedical sciences, economics, and the humanities remains largely untested.
Furthermore, many datasets exhibit acceptance bias. Authors and reviewers may often be reluctant to share rejected submissions because of reputational or institutional concerns. This imbalance skews label distributions and may distort score prediction models, increasing the risk of false positives in acceptance classification.

\textbf{Heterogeneous formatting and limited structure}. Peer review formats vary widely across venues. Some use structured templates, such as strengths and weaknesses, while others rely on free-form reviews. This inconsistency complicates dataset standardization and supervised training. Although pragmatic annotations and discourse labeling efforts exist~\cite{hua2019argument,dycke2023nlpeer}, they remain small-scale because of annotation costs. The lack of consistent structural metadata limits the ability to train models on fine-grained evaluative reasoning patterns.

\textbf{Legal and ethical barriers}. Many datasets lack explicit licensing terms governing reuse and redistribution. Reviewer anonymity, author consent, and institutional policies further constrain dataset sharing. These legal and ethical uncertainties inhibit the development of standardized, widely adopted benchmarks comparable to those in other NLP domains.

\subsection{Evaluation metrics}
Evaluating automated peer review systems is inherently challenging because review quality combines factual correctness, subjective judgment, structured reasoning, and constructive tone. Current evaluation protocols rely primarily on three approaches: automatic metrics, human assessment, and LLM-as-a-judge evaluation.

\textbf{Automatic metrics}. For score prediction and acceptance classification, discrete metrics such as accuracy, F1, mean absolute error (MAE)~\cite{willmott2005advantages}, and root mean square error (RMSE)~\cite{chai2014root} are commonly used. By contrast, critique generation is typically evaluated using text similarity metrics such as bilingual evaluation understudy (BLEU)~\cite{papineni2002bleu}, recall-oriented understudy for gisting evaluation (ROUGE)~\cite{lin2004rouge}, metric for evaluation of translation with explicit ordering (METEOR)~\cite{banerjee2005meteor}, BERTScore~\cite{zhang2019bertscore}, and MoverScore~\cite{zhao2019moverscore}(Table~\ref{tab:text_similarity_metrics}).
 However, these metrics were originally designed for tasks such as summarization or translation, not for evaluating critical, open-ended, and subjective peer reviews. As a result, they exhibit several limitations: 
 \begin{itemize}[nosep]
    \item Multi-reference ambiguity: Multiple valid reviews may exist for the same paper. Averaging scores across multiple human reviews can obscure complementary evaluations.
     \item Subjectivity mismatch: Peer reviews are inherently subjective, and rarely have a single correct critique. Metrics such as BLEU and ROUGE may penalize valid deviations from reference reviews. 
     \item Semantic insufficiency:  Embedding-based metrics capture semantic similarity more effectively than lexical metrics, but they still fail to assess constructiveness, factual correctness, or coherence of reasoning. 
     \item Reasoning blindness: Scientific reviews often involve structured reasoning, including critique, evidence, and suggestions; however, text similarity metrics generally ignore discourse structure or logical flow.
 \end{itemize}
 As a result, high ROUGE or BERTScore does not necessarily indicate an accurate, helpful, or insightful critique.

\textbf{Human evaluation}. Human evaluation remains the most reliable method for assessing helpfulness, constructiveness, factual correctness, and domain relevance. Studies commonly use Likert scales, pairwise comparisons, or domain-specific criteria such as novelty identification and flaw detection.
However, human evaluation faces substantial scalability challenges. It is time-intensive, costly, and subject to inter-rater disagreement, which mirrors the variability inherent in peer review itself. Consequently, human evaluation is often limited to small samples, reducing statistical power and making comparisons across systems difficult.

 \textbf{LLM-as-a-judge}. Recent studies have used LLMs to evaluate generated reviews through scoring or pairwise comparison. While scalable, this approach introduces risks of circularity: models trained on similar corpora may favor fluent but shallow critiques or outputs that resemble their own generation style. In addition, LLM-based judgments are sensitive to prompt design, model choice, and evaluation framing, and they often lack transparent reasoning mechanisms.
Thus, while promising, LLM-as-a-judge frameworks require further validation before they can serve as reliable benchmarks for automated peer review evaluation.

\begin{table*}[!t]
    \centering
    \small
        \renewcommand{\arraystretch}{1.3}
    \begin{tabular}{c|c|c}
    \hline
       Metric  & Evaluation &Limitations\\
       \hline 
       BLEU  & Precision-oriented n-gram overlap&penalize diversity, \\
       \cline{1-2} 
       ROUGE & Recall-based n-gram overlap &ignore semantic similarity\\
       \hline 
       METEOR&Synonym and stem matching + precision/recall&only consider shallow semantics\\
       \hline 
       BERTScore&Semantic similarity via contextual embeddings&fail to judge critique quality\\

&&Cannot verify factual correctness or consistency\\

       \cline{1-2}
       
       MoverScore &Embedding-based semantic alignment	&Often favor fluent text over insightful content\\
       \hline
    \end{tabular}
    \caption{List of well-known automatic text similarity metrics and their limitations in evaluating generated peer reviews.}
    \label{tab:text_similarity_metrics}
\end{table*}

\section{Security and Robustness Risks}
\label{security_risks}
As automated peer review systems move from experimental prototypes toward decision-support tools, robustness and security concerns become increasingly central. Unlike many conventional NLP tasks, peer review operates in a high-stakes setting in which evaluation outcomes can influence publication decisions, academic reputation, and career trajectories. As a result, vulnerabilities in LLM-based review systems may be subject to strategic exploitation. In this section, we outline potential risks across several paradigms for automated critique and scoring.
\subsection{Prompt injection and instruction manipulation}
Prompt-based review systems typically concatenate system instructions with manuscript content before generation. This architecture exposes them to prompt injection risks, whereby malicious or unintended instructions embedded in the manuscript influence model behavior. LLMs are known to be sensitive to prompt framing and contextual instructions~\cite{zhao2021calibrate}. If authors include hidden directives or strategically phrased content in their submissions, such as cues emphasizing novelty or instructions that redirect evaluative focus, models may inadvertently incorporate these signals into their critique or scores.

Recent work demonstrates the severity of this threat in scientific review settings. Keuper \textit{et al.}~\cite{keuper2025prompt} provide the first systematic analysis of such hidden manipulations, using 1,000 ICLR 2024 paper reviews across multiple LLMs. Their study shows that simple prompt injections can produce highly biased outputs, including artificially elevated acceptance scores, and in some cases can yield acceptance likelihood of up to 100\%. These findings reveal a concrete vulnerability in AI-assisted review workflows. Related evidence further suggests that prompt injections can induce topical shifts or alter review content when covert instructions are embedded within submissions~\cite{zhu2025your}. These results show that the manuscript itself becomes an attack surface in automated peer review pipelines: text that appears innocuous to human readers may contain embedded instructions that influence the evaluative behavior of an LLM.

Mitigating prompt injection risks requires robust input sanitization, including filtering invisible or embedded tokens, as well as adversarial robustness testing on curated corpora. Without such safeguards, structural vulnerabilities in LLM pipelines may be exploited, undermining both the reliability and integrity of automated peer review.
\subsection{Data poisoning in fine-tuned systems}
Fine-tuned review generation models derive their behavior directly from training data. Although supervised fine-tuning can improve structural alignment and task fidelity, it also exposes models to data poisoning and bias amplification risks. In the machine learning security literature, data poisoning refers to the intentional insertion of corrupted or malicious training examples that cause a learned model to behave erroneously on target inputs; such attacks have been demonstrated across classification, regression, and generation tasks~\cite{biggio2018wild}.

Fine-tuned peer review systems often rely on training corpora collected from open review platforms, such as OpenReview or PeerJ, as well as conference archives or volunteered peer reviews. This reliance introduces several risks:
\begin{itemize}[nosep]
    \item Adversarial poisoning: An adversary could deliberately inject strategically designed reviews into public platforms to skew the learned evaluation patterns of downstream models. For example, subtle labeling of weak submissions as “acceptable” or the systematic insertion of poorly justified positive critiques could bias models toward leniency.
    \item Bias amplification from noisy labels: Review data contains substantial label noise and subjective variation. Human reviewers often disagree, and scores typically reflect individual judgment rather than objective ground truth. When models are trained on such noisy and inconsistent labels, they may reproduce or amplify existing biases.
\end{itemize}
Data auditing, filtering, and bias assessment are therefore essential when constructing training corpora for automated peer review generation.

\subsection{Retrieval-based vulnerabilities}
Retrieval-augmented generation (RAG) systems reduce hallucination by conditioning outputs on retrieved evidence~\cite{lewis2020retrieval}. However, retrieval also introduces additional attack surfaces and robustness concerns. If external corpora contain outdated, misleading, or adversarially manipulated documents, generated critiques may incorporate inaccurate evidence. In novelty assessment tasks, reliance on incomplete literature indices may bias evaluation outcomes. Moreover, retrieval quality depends heavily on indexing strategies and embedding models, which may introduce topic imbalance or uneven coverage across research areas.
Robust evaluation of RAG-based review systems should therefore include retrieval accuracy metrics, corpus-quality checks, and stress tests under corpus perturbation.

\subsection{Reward hacking in alignment-optimized models}
Feedback-driven approaches have proven effective for aligning LLM outputs with human preferences. However, alignment objectives can introduce reward hacking risks, whereby models optimize for perceived helpfulness, politeness, or stylistic appeal rather than evaluative rigor. In peer review, reward functions that emphasize constructiveness or tone may inadvertently reduce critical sharpness. Over-optimization toward reward models may also produce outputs that maximize preference scores without improving factual correctness, depth, or calibration.
Balancing multi-objectives, including helpfulness, depth, grounding, and calibration, remains an open challenge in feedback-driven peer review systems.
\section{Deployment and Ethical Concerns}
\label{sec:challenges}
%While the previous section focused on the technical robustness and adversarial vulnerabilities of LLM-based review systems, 
This section addresses the governance, policy, and ethical considerations associated with deployment. These concerns extend beyond model reliability to include transparency, accountability, bias mitigation, privacy, and the appropriate role of AI in scholarly decision-making.
%The integration of LLMs into scientific peer review extends beyond questions of technical performance. Introducing automated critique and scoring systems often raises governance and ethical concerns that are distinct from modeling or robustness challenges. In this section, we examine deployment considerations for the responsible integration of LLM-based peer review systems.

\textbf{Accountability}. 
Traditional peer review operates within identifiable responsibility structures. Reviewers are selected to evaluate manuscripts, provide critiques, and recommend scores or acceptance decisions; chairs and editors then aggregate reviewers' feedback and make final decisions. When AI-generated outputs influence final decisions, responsibility becomes more diffuse. In particular, if an automated critique contributes to a rejection or acceptance outcome, several questions arise:
\begin{itemize}[nosep]
\item Who is responsible for potential errors or misleading assessments?
\item Should LLM-assisted reviews be explicitly labeled?
\item How can authors contest decisions influenced by LLM-generated content?
\end{itemize}

Scholarly publishing organizations have begun issuing guidance on the use of AI in peer review, emphasizing disclosure and reviewer accountability~\cite{ye2024we}. Maintaining clear human decision authority is therefore essential for preserving procedural fairness, accountability, and trust.

\textbf{Transparency and explainability}. 
Transparency is a central requirement in scientific evaluation. However, LLM-based systems often operate as black boxes with internal reasoning processes that remain opaque. Although explainable AI (XAI) methods aim to provide reasoning traces or confidence signals, generated explanations may not faithfully reflect underlying model computations~\cite{opusproject2024}. In the context of peer review, transparency and explainability span multiple dimensions:
\begin{itemize}[nosep]
    \item Disclosure of AI use by reviewers, chairs, or editorial systems.
    \item Documentation of model training data, system design, and intended use.
    \item Clear distinction between human-authored and AI-generated content.
    \item Reporting of uncertainty, calibration, and known limitations.
    \item Provision of grounded explanations, such as evidence or manuscript passages supporting generated critiques.
\end{itemize}
Without such mechanisms, opacity may undermine confidence in evaluation outcomes and hinder the responsible deployment of LLM peer review systems.

\textbf{Fairness}. LLM-based systems may inherit and amplify biases in historical review corpora, including biases related to institutional prestige, geography, gender, and disciplinary domains~\cite{hosseini2023fighting,pataranutaporn2025can}. Recent analyses suggest that LLM evaluators may favor fluent and formally structured manuscripts~\cite{ye2024we}, potentially disadvantaging non-native English authors or authors from less-resourced research environments. Moreover, if training datasets disproportionately reflect computer science venues or particular publication cultures, automated systems may encode discipline-specific norms as universal standards. Fair deployment, therefore, requires systematic auditing across author demographics, institutions, regions, and fields, as well as mechanisms for detecting and mitigating disparate impacts.

\textbf{Privacy and confidentiality}. Peer review involves unpublished research, proprietary methods, confidential findings, and in some domains, sensitive personal information. Manuscripts in areas such as biomedicine or social sciences may contain regulated or otherwise sensitive data. Submitting such materials to third-party LLM systems raises confidentiality risks~\cite{jin2024agentreview,ye2024we}, including data retention, logging, unauthorized reuse, or unintended exposure of novel findings or confidential data.
The deployment of AI-assisted review tools must therefore comply with institutional ethics requirements, publisher policies, and applicable data protection legal frameworks. At a minimum, systems should provide clear guarantees regarding data handling, retention, access control, and whether submitted manuscripts may be used for model training or system improvement.

\section{Application Scenarios}
\label{sec:application_scenarios}
Despite substantial challenges, LLM-based peer review systems offer promising opportunities when deployed responsibly. Rather than viewing automation as a binary replacement of human reviewers, it is more productive to consider a spectrum of application scenarios, ranging from author assistance to editorial augmentation. In this section, we outline practical pathways for deploying automated peer review systems.

\textbf{Pre-submission author assistance}. One of the most immediate and relatively low-risk applications of LLM-based review systems lies in the pre-submission stage. Authors can use LLM-generated critiques to assess the clarity, structure, methodological soundness, and novelty framing of their manuscripts before formal submission. Recent studies suggest that LLMs can produce structured and useful feedback that approximates review-style commentary, although their depth, calibration, and reliability remain imperfect~\cite{liang2024can,idahl2025openreviewer}. In this setting, LLM systems may serve as automated writing and critique assistants, helping authors identify ambiguous claims, missing experimental details, inconsistencies, or weakly supported conclusions. Such tools could reduce avoidable errors and improve manuscript quality before peer review. However, deployment in this context requires careful attention to confidentiality, since uploading unpublished work to external services may expose intellectual property. Institutionally hosted systems or locally deployed models may therefore provide safer alternatives for pre-submission use.

\textbf{LLMs as first-pass screening}. Another potential application is first-pass screening. Conferences and journals often receive a large volume of submissions, creating substantial burdens for editors, area chairs, and reviewers. LLM-based systems could provide preliminary signals for editorial triage, such as identifying submissions that may require additional scrutiny, missing required components, or unclear methodological claims. However, they should not be used as stand-alone mechanisms for desk rejection or acceptance decisions. 
Such systems could also support iterative improvement by offering authors early feedback before human review. For example, authors might respond to LLM-generated comments or revise their manuscripts before subsequent evaluation, thereby reducing avoidable issues and improving the efficiency of later review stages. A two-stage pipeline of this kind could improve scalability while preserving human oversight for final editorial and acceptance decisions.

\textbf{LLMs as co-reviewer}. LLM-generated reviews may also be used alongside human reviews to provide complementary perspectives. In this role, automated reviews would not replace human judgment but would instead serve as an additional source of feedback, potentially highlighting issues that human reviewers overlook or offering alternative interpretations of a manuscript’s contributions and limitations. For example, AAAI 2025 launched a pilot program in which LLM-generated reviews were shared with authors as supplementary feedback, enriching the evaluation process with additional perspectives~\cite{aaai}. Such deployments may be particularly useful when clearly labeled, carefully calibrated, and separated from binding decision-making authority.

\textbf{LLMs as reviewer assistance}. LLMs can also support reviewers through auxiliary tasks such as summarizing manuscripts, verifying citations, and checking internal consistency, identifying missing experimental details, and improving the clarity of review comments. These functions may allow human reviewers to focus more directly on substantive evaluation and final judgment~\cite{luo2025llm4sr}. For instance, a recent ICLR pilot experiment used LLMs to flag inappropriate remarks, highlight manuscript sections relevant to reviewers' questions, and suggest clearer phrasing for vague review comments~\cite{iclr}. Beyond these auxiliary uses, LLMs could also generate initial review drafts based on manuscript content, which human reviewers would then revise, contextualize, and personalize by adding their own evaluations and insights. However, over-reliance on AI-generated content remains a significant risk. To mitigate this concern, reviewers should retain responsibility for the final review and may be required to justify cases in which their final assessments closely align with LLM-generated suggestions.
\section{Future research directions}
\label{future_work}
Despite rapid advances in LLM-based critique and score generation, automated scientific peer review remains an emerging research area. Future work should move from review-style generation toward structured evaluative modeling that tests novelty, evidence support, methodological rigor, claim consistency, and the relationship between contributions and prior literature. Promising directions include decomposing review generation into interpretable sub-tasks, incorporating tool-augmented reasoning, and leveraging structured representations such as argument graphs or claim–evidence mappings to better capture the underlying processes of scientific critique.
%Future work should move from review-style generation toward structured evaluative modeling that tests novelty, evidence support, methodological rigor, and claim consistency. Scientific evaluation requires deeper forms of reasoning, including assessing novelty, evaluating methodological rigor, verifying consistency between claims and evidence, and contextualizing contributions within the broader literature. Future research should therefore move beyond surface-level generation toward structured evaluative modeling.

Another important direction is to expand beyond the current computer science–centric datasets. Future benchmarks should test whether models trained on ML/NLP review norms transfer to fields with different evidentiary standards, such as biomedicine, economics, and the humanities. Future work should explore domain adaptation techniques and develop more domain-diverse benchmarks that reflect the norms, evaluation criteria, and evidentiary standards of different scientific communities.

%Robustness and security also remain underexplored but increasingly critical. 
Robustness research should move from general warnings to benchmarked stress tests, including prompt-injection suites, poisoned-review training sets, retrieval-corruption tests, and adversarial manuscripts. Such benchmarks should evaluate not only whether attacks change generated text, but also whether they distort scores, alter critique emphasis, or affect downstream editorial decisions.
%Further research on adversarial stress testing, secure retrieval design, and robustness-aware benchmarking will be important for strengthening the reliability and trustworthiness of LLM-based peer review systems.

Finally, richer evaluation paradigms are needed to better align automated metrics with the goals of scientific critique. Traditional text similarity metrics are insufficient for capturing reasoning depth, critique coverage, factual grounding, and logical coherence. Future evaluation should prioritize evidence-based assessment across multiple reviews, as well as human-AI agreement modeling that goes beyond simple correlation-based measures.
\section{Conclusion}
\label{sec:conclusion}
This study examined the use of LLMs for automated peer review generation across several key dimensions, including modeling approaches, benchmark resources, security risks, deployment considerations, and practical application scenarios. Our analysis highlights that although LLMs can produce fluent, well-structured reviews, substantial challenges remain in calibrated scoring, deep methodological reasoning, cross-domain generalization, robustness, and fairness. Moreover, the integration of automated systems into peer review introduces broader institutional concerns related to transparency, accountability, and confidentiality.

Despite these limitations, LLMs hold considerable promise as assistive tools in the scholarly review process. They can support authors in pre-submission refinement, help reviewers with summarization and consistency checking, and assist editors and program chairs in managing review workflow. When appropriately designed and deployed, such systems may reduce reviewer burden and improve workflow efficiency. The most promising near-term direction is a hybrid human–AI collaboration model that preserves human decision authority while leveraging the scalability of automated assistance. Looking ahead, progress in automated peer review will require advances in evaluative reasoning, task-specific benchmarking, robustness testing, transparency mechanisms, and fairness-aware system design. 
The central challenge is therefore not whether LLMs can imitate the surface form of peer review, but whether they can be integrated as calibrated, transparent, and auditable decision-support tools within human-led evaluation workflows.
%With careful development and responsible integration, LLM-based systems have the potential to strengthen the integrity of scientific peer review.

\bibliographystyle{abbrv}
\bibliography{references}

\end{document}